\documentclass[conference]{IEEEtran}
\IEEEoverridecommandlockouts
\usepackage{cite}
\usepackage{amsmath,amssymb,amsfonts}
\usepackage{algorithmic}
\usepackage{graphicx}
\usepackage{textcomp}
\usepackage{xcolor}
\def\BibTeX{{\rm B\kern-.05em{\sc i\kern-.025em b}\kern-.08em
    T\kern-.1667em\lower.7ex\hbox{E}\kern-.125emX}}
\usepackage{soul}
\usepackage{booktabs} % for professional tables
\usepackage[flushleft]{threeparttable} % for notes below table
\usepackage{tabu}
\usepackage{extarrows}
\usepackage{xspace}
\usepackage{lipsum}
\usepackage{nicefrac}
\usepackage{pifont}
\newcommand{\cmark}{\ding{51}}%
\newcommand{\xmark}{\ding{55}}%
\usepackage{textgreek} % for uLayer reference from EuroSys 2019.
\usepackage{background}

\newcommand{\inlineeqnum}{\refstepcounter{equation}~~\mbox{(\theequation)}}

\makeatletter
\def\footnoterule{\relax%
  \kern-5pt
  \hbox to \columnwidth{\hfill\vrule width 1\columnwidth height 0.4pt\hfill}
  \kern4.6pt}
\makeatother

\newif\ifcomment

\commenttrue

\commentfalse 

\ifcomment
\definecolor{stelios_colour}{RGB}{144, 238, 144}
\newcommand{\stelios}[1]{\sethlcolor{stelios_colour}\hl{[\textbf{Stelios:} #1]}}
\definecolor{giannis_colour}{RGB}{191, 232, 255}
\newcommand{\giannis}[1]{\sethlcolor{giannis_colour}\hl{[\textbf{Giannis:} #1]}}
\else
\newcommand{\stelios}[1]{}
\newcommand{\giannis}[1]{}
\fi

\newcommand{\tool}{OODIn\xspace}

\renewcommand{\IEEEbibitemsep}{0pt plus 0.5pt}
\makeatletter
\IEEEtriggercmd{\reset@font\normalfont\fontsize{7.9pt}{8.40pt}\selectfont}
\makeatother
\IEEEtriggeratref{1}

\backgroundsetup{angle=0,
    scale=1,
    color=black,
    firstpage=true,
    position=current page.north,
    hshift=0pt,
    vshift=-20pt,
    contents={\ifnum\value{page}=1 PREPRINT: Accepted at the 7th IEEE International Conference on Smart Computing (SMARTCOMP), 2021 \else \fi}
}

\begin{document}

%\title{OODIn: Optimised On-Device Inference}
% \title{OODIn: A Deep Learning Framework for Optimised On-Device Inference}
\title{\vspace{-0.4cm} OODIn: An Optimised On-Device Inference Framework for Heterogeneous Mobile Devices \vspace{-0.4cm}}

% \vspace{-0.7cm}}

\author{\IEEEauthorblockN{
Stylianos I. Venieris\IEEEauthorrefmark{2},   
Ioannis Panopoulos\IEEEauthorrefmark{3},
Iakovos S. Venieris\IEEEauthorrefmark{3}
}
% \\
\IEEEauthorblockA{\IEEEauthorrefmark{2}Samsung AI Center, Cambridge, UK,
\IEEEauthorrefmark{3}National Technical University of Athens, Athens, Greece}
% \IEEEauthorblockA{Email: \{s.venieris\}@samsung.com, ioannispanop@mail.ntua.gr, venieris@cs.ece.ntua.gr}
\vspace{-0.8cm}
}

\maketitle

\begin{abstract}
% Random text as placeholder

Radical progress in the field of deep learning (DL) has led to unprecedented accuracy in diverse inference tasks. 
%, from object %and speech 
% recognition to 
%scene understanding and 
% health monitoring. 
As such, deploying DL models across mobile platforms is vital to enable the development and broad availability of the next-generation intelligent apps. Nevertheless, the wide and optimised deployment of DL models is currently hindered by the vast system heterogeneity of mobile devices, the varying computational cost of different DL models and the variability of performance needs across DL applications.
This paper proposes \textit{\tool}, a framework for the optimised deployment of DL apps across heterogeneous mobile devices. \tool comprises a novel DL-specific software architecture together with an analytical framework for modelling DL applications that: (1)~counteract the variability in device resources and DL models by means of a highly parametrised multi-layer design; and (2)~perform a principled optimisation of both model- and system-level parameters through a multi-objective formulation, designed for DL inference apps, in order to adapt the deployment to the user-specified performance requirements and device capabilities. Quantitative evaluation shows that the proposed framework consistently outperforms status-quo designs across heterogeneous devices and delivers up to 4.3$\times$ and 3.5$\times$ performance gain over highly optimised platform- and model-aware designs respectively, while effectively adapting execution to dynamic changes in resource availability.
\end{abstract}

\vspace{-0.1cm}
\section{Introduction}
\label{sec:intro}
\vspace{-0.1cm}

In recent years, deep learning (DL) models have emerged as the state-of-the-art in several AI inference tasks. Ranging from the recognition of objects~\cite{resnet2016cvpr} and emotions~\cite{emotion_recognition2021nature} to scene~\cite{deeplabv32017arxiv} and speech understanding~\cite{ondevice_asr2020getmobile}, their unrivalled accuracy has made deep neural networks (DNNs) an enabler of many mobile apps. 
Currently, developers who seek state-of-the-art accuracy and wide device compatibility typically resort to offloading the DL model execution to a remote server~\cite{neurosurgeon2017asplos}. While this approach can resolve the problem of supporting devices with different capabilities, cloud/edge offloading comes with high operating costs, raises privacy concerns due to the transmission of user data and leads to inconsistent user experience due to varying networking conditions when using cellular connectivity.

Driven by recent developments in both the hardware and algorithmic fronts, \textit{on-device} execution is becoming a promising alternative. On the one hand, device vendors have started enhancing their chipsets with specialised DNN processors, often called Neural Processing Units (NPUs)~\cite{ai_benchmark2019iccvw}. On the other hand, numerous model compression techniques have led to significantly more compact DNNs~\cite{dnn_eff_proc2017procieee}, enabling their deployment on resource-constrained platforms. As a result, systems that rely only on the local resources of the user device to execute DNNs are increasingly more competitive.

Despite the progress, mainstream mobile usage of DNNs is primarily isolated to only a few global-scale companies that have the human and computational resources to build proprietary solutions. This can be attributed to three main factors: \textit{1)}~the diversity in the processing capabilities of devices in the wild, leading to wide system heterogeneity and inconsistent performance across devices; \textit{2)}~the variety of DL models, in terms of task, architecture and resource demands; and \textit{3)}~the variability of performance requirements in terms of accuracy, latency, throughput and energy across DNN applications.
This situation is further aggravated by the fluctuation of resource availability due to the multi-tasking nature of smartphones.

In this work, we propose \textbf{\texttt{\tool}}, a framework that makes strides towards overcoming the barriers that hinder the wide integration of DL in mobile devices.
\tool introduces a highly customisable mobile software architecture that allows the manipulation of critical model- and system-level parameters in order to adapt to both the DNN workload and the device at hand.
Moreover, by means of a multi-objective modelling framework, \tool closely captures the various performance demands of DL applications and tailors execution to the given use-case.
The key contributions of this paper are the following:
\begin{itemize}
    \item A novel software architecture that enables running a broad range of DL applications across mobile devices while meeting stringent performance constraints. 
    The proposed architecture introduces a highly parametrised multi-layer design that provides the necessary building blocks for developing smart applications, while carefully decoupling the application functionality from the selection of DL model and the specifics of the device resources.
    Moreover, through its set of tunable parameters, the developed architecture introduces fine-grain customisability at both the model and system level, aiming to attain the maximum device capabilities and satisfy the user-defined performance targets.
    
    \item The \tool automated framework for deploying DL applications across diverse mobile devices. The developed framework first takes as input a target DNN in \mbox{TensorFlow} together with multiple performance objectives. Leveraging our multi-objective optimisation formulation, \tool traverses the design space across both model- and system-level parameters and yields a highly optimised configuration, tailored to the various application requirements and the characteristics of the target device. Upon deployment, \tool tracks changes in system resource usage and reconfigures its parameters accordingly through a run-time adaptation mechanism.
    
    \item We implement \tool on commodity smartphones and conduct comprehensive experiments to evaluate its effectiveness in deploying diverse DL applications across heterogeneous devices, without compromising performance.
\end{itemize}

%\vspace{-0.1cm}
\section{Deep Learning on Mobile Devices}
\label{sec:background}
% \vspace{-0.1cm}

\begin{figure*}[t!]
    \centering
    \vspace{-0.7cm}
    \includegraphics[width=0.8\textwidth]{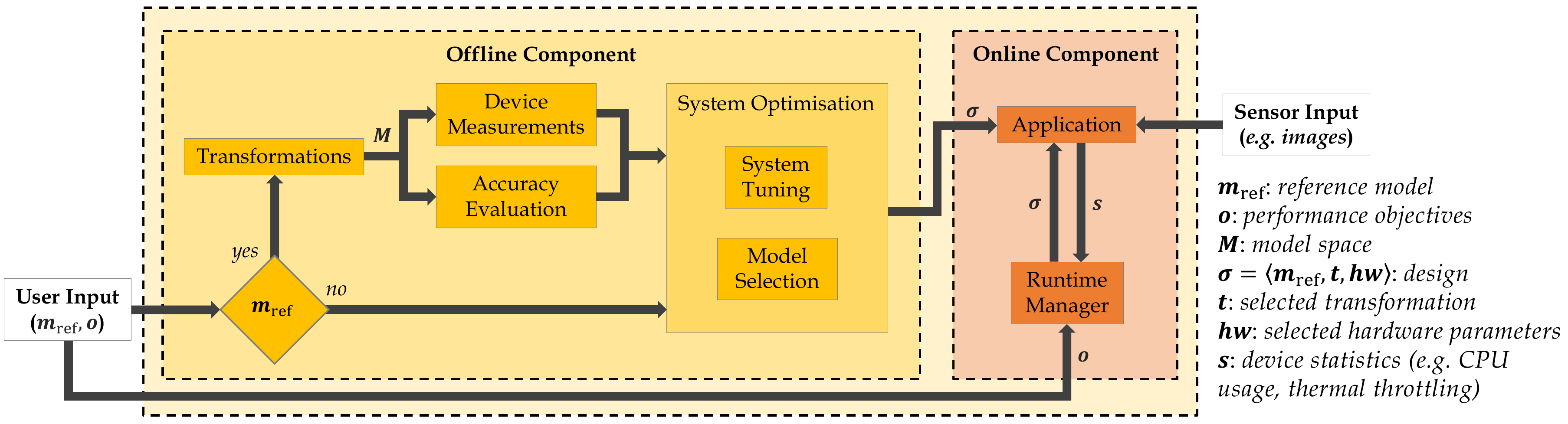}
    \vspace{-0.2cm}
    \caption{Overview of \tool's processing flow.}
    \vspace{-0.4cm}
    \label{fig:processing_flow}
\end{figure*}

\textbf{DL Model Compression.} Several techniques have been proposed for simplifying DNNs~\cite{dnn_eff_proc2017procieee} in order to match the capabilities of user devices, such as resource-limited smartphones and IoT modules.  
Methods such as pruning~\cite{netadapt2018eccv}, quantisation~\cite{integer_dnn2018cvpr} and early-exit models~\cite{hapi2020iccad} all aim to reduce the size, latency and energy consumption of a DNN.
Nevertheless, with the majority of techniques developed as proprietary solutions, their integration into mobile apps remains challenging and their adoption in real-world DL apps limited~\cite{dl_smartphones2019www}.
Moreover, the attainable computational savings vary significantly across heterogeneous devices. 
Thus, there is an emerging need for algorithmic and system solutions that provide a principled adaptation of such techniques to diverse devices and DNNs.

\textbf{DL Backends for Mobile.}
Several frameworks are gaining traction for mobile inference. Two prominent and widely used frameworks are TensorFlow Lite (TFLite)\footnote{https://tensorflow.org/lite} and PyTorch Mobile.\footnote{https://pytorch.org/mobile/home/}
In addition to converting trained models into a format suitable for on-device inference, TFLite can also optionally optimise DNNs with various compression methods, such as quantisation or weight clustering, while also supporting multiple processor targets, such as GPUs and NPUs via Android NNAPI~\cite{ai_benchmark2019iccvw}. 
Nonetheless, both frameworks put emphasis solely on the efficient execution of DNN inference and do not capture critical application-level needs, such as multiple -and often competing- performance objectives. Moreover, most optimisations are targeted only towards flagship devices, leaving the rest of the device landscape with limited to no support~\cite{fb_edge2019hpca}.

\textbf{Challenges of Mobile DL.}
The methods and tools mentioned above are confronted with the great heterogeneity found in mobile devices, which stems from their different System-on-Chips (SoCs)~\cite{ai_benchmark2019iccvw}. In other words, there is no ``typical" smartphone or SoC and this makes it unusually difficult to apply optimisations that generalise across diverse devices~\cite{fb_edge2019hpca}. 
At the same time, AI applications come with various performance needs. These span from latency-critical apps, such as smart cameras and Augmented Reality (AR)~\cite{mobileAR2019mobicom}, to the throughput-oriented demands of high-resolution video understanding~\cite{deeplabv32017arxiv} and image enhancement~\cite{neural_enhancement2021csur}.
Finally, DL models tend to have vastly different computational costs. For instance, InceptionV3 requires an order of magnitude more FLOPs and memory than EfficietNetLite0 (Table~\ref{tab:dnns}).
The situation is further aggravated by the limited available resources of mobile devices compared to cloud/edge-based systems.
All this makes it difficult to develop end-to-end applications that provide %flexibility and 
interoperability between devices while sustaining high performance.

\vspace{-0.1cm}
\section{\tool}
\label{sec:system}
\vspace{-0.1cm}

\vspace{-1mm}
\subsection{Overview}
\label{sec:overview}
\vspace{-1mm}

To bridge the gap between DL and mobile, \tool is proposed (Fig.~\ref{fig:processing_flow}).
\tool addresses the main challenges of mobile DL (\S\ref{sec:background}) at three levels. 
First, a novel DL application model is introduced that captures DL applications by means of a \textit{multi-objective optimisation framework} (Section~\ref{sec:optimisation}). This enables us to represent various use-cases along the most important performance dimensions and analytically express their relative importance for the use-case at hand. 
Second, to tunably configure the accuracy-complexity trade-off of a given DL model and adapt it to the target device capabilities, a novel model selection \giannis{italics?} method is employed which ensures that most performance requirements will be met regardless of the device (Section~\ref{sec:offline_component}). Finally, to counteract system heterogeneity, we introduce the \textit{Mobile Device Convergence Layer} (MDCL) (Section~\ref{sec:cl}), a thin device-aware software wrapper that ensures portability and scalability of DL applications across mobile platforms. This is achieved by abstracting the resource details of the underlying device and exposing critical system parameters that can be optimised, either offline or at run time, to extract the desired performance from the hardware.
Overall, the following design goals were taken into consideration.

\noindent
\textbf{Flexibility:} 
Driven by the robust accuracy of DL models across domains, \tool should be able to optimise a wide range of use-cases that arise through modern mobile apps. This includes enabling various tasks, while satisfying the multiple performance objectives of emerging applications.

\noindent
\textbf{Model Independence:}
Given the large variety of existing DNNs, \tool is required to support and provide optimised execution of DL models with different architectures, number of layers and overall computational and memory footprint.

\noindent
\textbf{Device Portability:}
Device independence entails the system software and the underlying hardware. Given the fragmented landscape of OS variants, 
\tool needs to adapt apps 
such that they sustain their functionality and performance. On the hardware front, the existence of multiple vendors has led to devices with significantly different resource characteristics, including camera, screen, memory and processors. \tool needs to consider the device-specific resources and accordingly customise the mapping of the target app.

\noindent
\textbf{Scalability:} 
With the mobile devices' capabilities varying significantly, there is a need for tools that abstract the resource details of a particular device and provide scalability.
\tool should be able to sustain or improve performance in case of an increase in the amount of available resources.

\noindent
\textbf{Adaptability:}
Given the intrinsic dynamicity of mobile apps due to multi-tasking, which in turn affects the instantaneous availability of resources, \tool should tunably balance the trade-off between performance and resource usage in order to adapt to dynamic changes at run time.

\subsection{Processing Flow}
\label{sec:proc_flow}

Fig.~\ref{fig:processing_flow} depicts \tool's flow, consisting of two stages: the offline and the online component.
As a first step, the user provides the performance objectives and optionally their own custom DL model. The offline component starts by applying a set of \textit{Transformations} over the supplied model in order to derive a number of compressed model variants. To assess their characteristics, each variant is evaluated with respect to its accuracy and resource demands. \textit{Accuracy Evaluation} is conducted using a user-supplied validation dataset for the task at hand. On-device performance is measured through a number of \textit{Device Measurements} on a given mobile device for each model. These measurements monitor latency, throughput and memory usage and are repeated for varying system-level parameters. 
Next, the accuracy and performance results are passed to the \textit{System Optimisation} module. 
At this stage, \tool traverses the design space defined by the different model variants and system-level parameters, in order to yield the highest performing configuration that satisfies the user's performance objectives. 
Finally, the offline component's selected configuration is loaded into the \textit{Application}. At run time, the \textit{Runtime Manager} monitors the instantaneous resource usage and adapts the app configuration to dynamic changes.

\subsubsection{Offline Component}
\label{sec:offline_component}
Internally, \tool represents a given model as a tuple $m = \left<task, w, s_m, s_\text{in}, a, p \right>$, where $task$ is the DL task (\textit{e.g.}~object detection), $w$ is the workload in number of FLOPs, $s_m$ is the model size (\textit{i.e.}~number of parameters), $s_{\text{in}}$ is the resolution of the input samples, $a$ is the accuracy and $p$ is the numerical precision.

\textbf{Model Space.}
The user-supplied DL model constitutes the starting point for \textit{Model Selection} (Fig.~\ref{fig:processing_flow}). The space of model variants is defined based on \textit{Transformations} employed by our framework in order to modify the accuracy-complexity trade-off of the model. Given a transformation set $\mathcal{T}$, we generate a new model variant by applying one transformation from $\mathcal{T}$ on the original model $m_{\text{ref}}$ as $m \xlongleftarrow{t} m_{\text{ref}}$
for $t \in \mathcal{T} \inlineeqnum \label{eq:transform}$.

This formulation spans a model space which we denote by {\small $\mathcal{M} = \left\{ m ~ | ~ m \xlongleftarrow{t} m_{\text{ref}}, ~ \forall t \in \mathcal{T} \right\}$}. 
Without hurting the generality of our framework, the set $\mathcal{T}$ currently contains various post-training quantisation schemes, including half-precision floating-point (FP16), 8-bit fixed-point (INT8) and the conventional full-precision floating-point (FP32), leading to $\mathcal{T}$$=$$\{\text{FP32}, \text{FP16}, \text{INT8} \}$. As such, $\mathcal{T}$ can be extended to include other techniques that trade off accuracy and complexity, such as pruning~\cite{netadapt2018eccv} or dynamic channel skipping~\cite{nestdnn2018mobicom}.

\textbf{System Parametrisation.}
To enable fine-grain customisation, \tool introduces system-level parameters that can be tuned to tailor the execution of the DNN to both the performance needs and the underlying hardware.
First, we represent the available resources on the target platform as 
\begin{equation}
    \small
    R = \left<\mathcal{CE}, N_{\text{cores}}, C, \mathcal{DVFS}, b, v_{\text{os}}, v_{\text{camera}} \right>
    \label{eq:target_platform}
\end{equation}
where $\mathcal{CE}$ is the set of available compute engines, $N_\text{cores}$ the number of CPU cores, $C$ the memory capacity, $\mathcal{DVFS}$ the set of available Dynamic Voltage and Frequencey Scaling (DVFS) governors, $b$ the battery capacity, $v_{\text{os}}$ the version of the Android OS and $v_{\text{camera}}$ captures the camera characteristics, such as the available APIs, the screen resolution and the flash capabilities.

Given this representation, we introduce the following tunable system parameters: \textit{1)}~the task-to-processor mapping, which selects which compute engine $ce$$\in$$ \mathcal{CE}$ will perform the inference computations; \textit{2)}~the number of threads $N_{\text{threads}}$$\in$$\{1, ..., N_{\text{cores}}\}$ when using multithreading on the CPU; \textit{3)}~the governor $g \in \mathcal{DVFS}$ which determines the DVFS policy of the device; and \textit{4)}~the recognition rate $r$, which determines the invocation frequency of the inference engine, \textit{e.g.}~when $r=1$, inference is performed on each frame, while when $r=0.5$, every second frame. Overall, the configurable system-level parameters of \tool are $hw$$=$$\left<ce, N_{\text{threads}}, g, r \right>$.

\textbf{System Optimisation.}
\tool's strategy of selecting both the most suitable model variant (\textit{Model Selection}) and system parameters (\textit{System Tuning}) for each application is based on the performance objectives, the generated DL models and the device capabilities. The \textit{System Optimisation} module navigates the space of candidate designs across both the model and system dimensions and yields the most suitable design $\sigma = \left<m_{\text{ref}}, t, hw \right>$. \textit{System Optimisation} is discussed in detail in Section~\ref{sec:optimisation}. 
In case the user has not supplied \tool with a custom reference model, our framework suggests the best fit from the set of available models shown in Table~\ref{tab:dnns}.

\subsubsection{Online Component}
\label{sec:online_component}
The online component of \tool consists of the mobile \textit{Application} and the \textit{Runtime Manager}. The selected design $\sigma$ is used to configure the \textit{Application}, which can then be launched by the user.
Upon deployment, the \textit{Application} continuously monitors the system resources and sends periodic statistics about the state of the device, such as the cores' temperature or memory usage, to the \textit{Runtime Manager}. Based on this information, the \textit{Runtime Manager} decides whether to modify the system parameters $hw$ or even the model itself through a different transformation $t$, in order to adapt to any significant changes in resource availability.

\subsection{Mobile Software Architecture}
\label{sec:sw_arch}
\begin{figure}[t]
    \vspace{-0.2cm}
    \centerline{\includegraphics[width=0.49\textwidth]{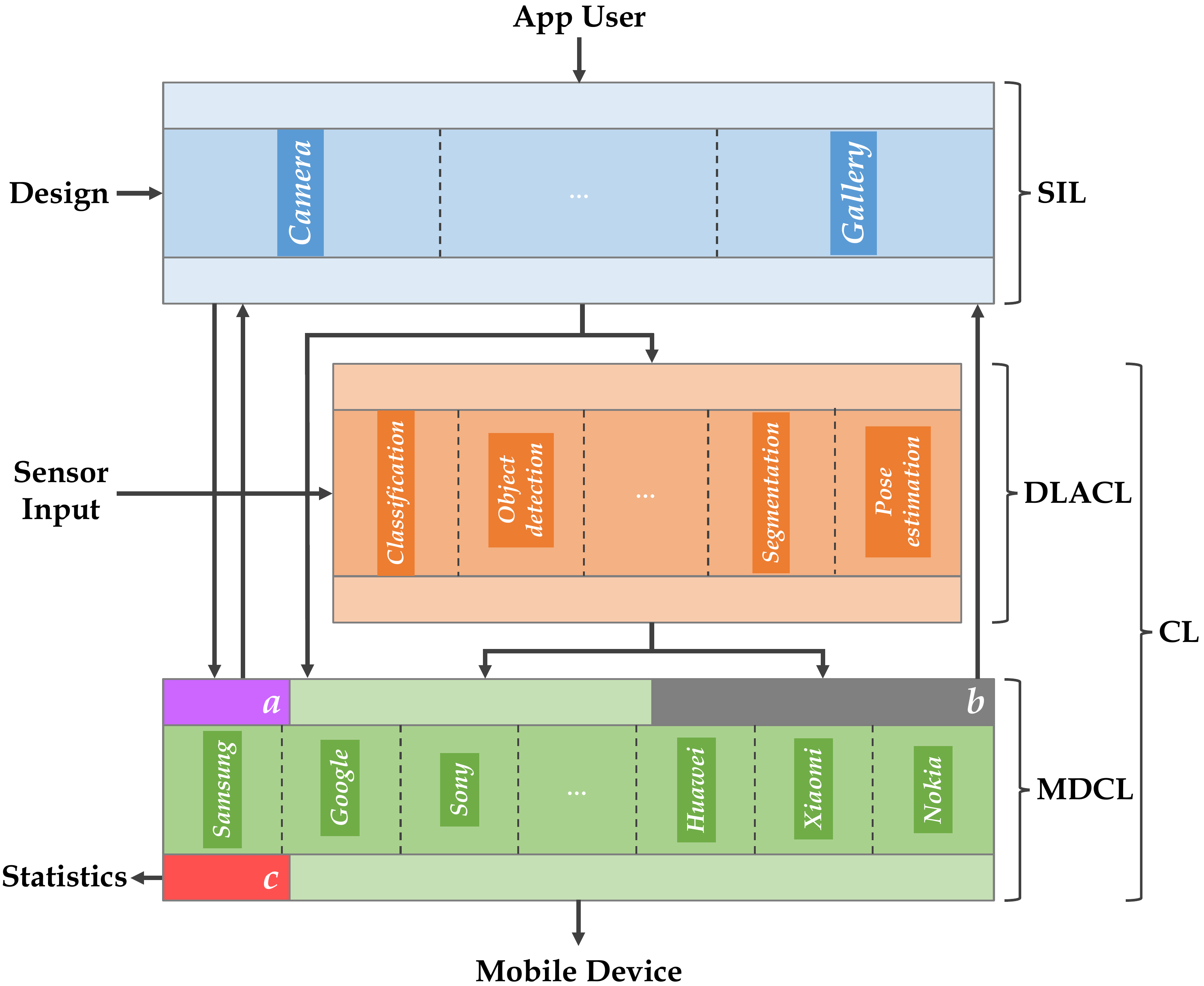}}
    \vspace{-0.2cm}
    \caption{\tool's multi-layered software architecture.}
    \vspace{-0.4cm}
    \label{fig:sw_arch}
\end{figure}

Given the design considerations  in \S\ref{sec:overview}, we adopt a multi-layer software architecture for \tool's mobile \mbox{\textit{Application}} component. 
As depicted in Fig.~\ref{fig:sw_arch}, \tool's architecture consists of two coarse layers: \textit{1)}~the \textbf{S}ervice-\textbf{I}ndependent \textbf{L}ayer (SIL) and \textit{2)}~the \textbf{C}onvergence \textbf{L}ayer (CL).

\subsubsection{Service-Independent Layer}
\label{sec:sil}
The primary objective of SIL is to provide the app-level functionality while abstracting the hardware resource details of the target device and the characteristics of the DNN model.
To this end, SIL is designed to supply the core building blocks for constructing various smart applications.
As such, SIL offers a selection of widely used modules, such as a camera interface for real-time visual apps, a local database for storing processed data, \textit{e.g.}~the \tool-labelled photos of a user in a smart Gallery app, and user interface (UI) components for interacting with the user.
SIL packages the aforementioned blocks under a unified API to allow developers to combine them in a flexible, modular and maintainable manner.
After SIL completes the configuration of the app and the UI, the \textit{Runtime Manager} as well as the inference engine for the DL model can be initialised.

\subsubsection{Convergence Layer}
\label{sec:cl}
In a complementary manner to the model- and platform-agnostic SIL, CL considers the DNN model and the system specifications of the given device, and performs a number of adaptation steps. Thus, we organise CL into two sublayers: the \textbf{D}eep \textbf{L}earning \textbf{A}rchitecture \textbf{C}onvergence \textbf{L}ayer (DLACL) which considers the given DNN model; and the \textbf{M}obile \textbf{D}evice \textbf{C}onvergence \textbf{L}ayer (MDCL) whose operation considers the target device. Both layers are parametrised in order to provide the necessary fine-grain customisability that will offer the flexibility to support the diverse performance needs of DL applications.

\textbf{DLACL.}
This layer provides the first DNN-aware interface, capturing information about the DL architecture and its resource demands. DLACL is highly decoupled from both SIL and MDCL as it is mainly responsible for receiving incoming data samples from the input source (\textit{e.g.}~camera, storage or microphone) and feeding them to the inference engine.

In addition, DLACL implements the online model selection, whenever the \textit{Runtime Manager} dictates that a different model variant should be used (\S\ref{sec:online_component}). To alternate between the available models, this layer needs the ability to dynamically allocate sufficient memory for each model. To this end, we isolate the set of buffers that are model-dependent, so that they are managed only by DLACL. This includes buffers for the input samples, the model itself and intermediate results. The buffer sizes rely on the size of the inputs ($s_{\text{in}}$), the model size ($s_m$) and the precision ($p$), \textit{i.e.}~the number of bits used to represent the model's trained parameters, respectively. Hence, DLACL internally adopts the same tuple representation as \tool's offline component (\S\ref{sec:offline_component}) and stores $s_{\text{in}}$, $s_m$ and $p$ for each model. As these values are known \textit{a priori} before deployment, this approach allows DLACL to statically determine how much memory is needed for each buffer and, thus, allocate only the necessary amount, without starving the memory resources whenever a model swap is performed. 

\textbf{MDCL.}
Further closer to the hardware, the MDCL layer is responsible for identifying the resources of the target platform and allocating them to the task at hand based on its needs. As such, MDCL populates the target platform resource model $R$~(Eq.~(\ref{eq:target_platform})).
For instance, the representation of Samsung S20 FE is constructed as $\mathcal{CE}_{\text{S20FE}}$$=$$\{CPU, GPU, NPU\}$, $N_{\text{cores}}$$=$$8$, $C$$=$$6$ GB, $\mathcal{DVFS}$$=$$\{\scriptsize \texttt{energy\_step}, \texttt{performance}, \texttt{schedutil}\}$, $b$$=$$4500$~mAh, $v_{\text{os}}$$=$$11$ and $v_{\text{camera}}$$=$$\{\texttt{FULL}, ..., 1080\times2400\}$.
The detected parameters are also used when performing \textit{Device Measurements} (Fig.~\ref{fig:processing_flow}) in order to sweep over valid value ranges, \textit{e.g.}~benchmark each model variant's execution on all available compute engines in $\mathcal{CE}$ or varying the threads up to $N_{\text{cores}}$.

MDCL is equipped with three independent middlewares (Fig.~\ref{fig:sw_arch}). Middleware $a$
provides SIL with the necessary hardware information, which is vital for the configuration of the application's basic components, \textit{e.g.}~the camera interface depends on the type of visual sensors and the UI on the screen resolution. Middleware $b$ is optionally used for optimising the application's features based on the output of DLACL. For instance, an AI Camera app could optionally optimise the parameters of the camera (\textit{e.g.}~the brightness) based on the output of a scene recognition DNN for the previous frame. Middleware $c$ is responsible for collecting and transferring various system statistics, such as GPU or memory usage to the \textit{Runtime Manager}. This middleware can also send warnings regarding unexpected behaviour, such as CPU throttling.

\subsection{System Optimisation}
\label{sec:optimisation}

The developed framework aims to determine a model transformation $t$ together with the values of system parameters $hw$ that optimise the user-defined performance objectives. We denote a candidate configuration of \tool as $\sigma = \left<m_{\text{ref}}, t, hw \right>$.

To provide a flexible modelling framework for DL applications, we adopt a multi-objective optimisation (MOO) formulation. Under this scheme, we cast each DL application as a MOO problem based on the performance metrics of interest. The current set of performance metrics is defined as \mbox{$\mathcal{P}=\{T, fps, mem, a\}$} where $T$ is the latency, $fps$ the throughput in frames-per-second, $mem$ the memory footprint and $a$ the DNN accuracy. For each of the metrics, the user can define whether it should be maximised or minimised, or whether the average, median or $n$\textsuperscript{th} percentile should be as close as possible to a target value $val$.
Formally, we define the $i$-th user-specified objective as {\small $o_i$$=$$\left<P, max/min/val(avg/median/n^{\text{th}}) \right>$} where metric $P \in \mathcal{P}$ is optimised as specified by the second element of the tuple.  
We present three representative use-cases in Eq.~(\ref{eq:max_fps})-(\ref{eq:max_acc_max_fps}).

%\noindent
\textbf{Use-case 1. MaxFPS:}
Optimise the application to achieve the maximum throughput in frames-per-second (FPS) without degrading a given level of accuracy.
% \vspace{-1mm}
\begin{equation}
    \small
    \max\limits_{\sigma} fps(\sigma) \quad \text{s.t.} \quad a_{1}(\sigma) - a_{1,\text{ref}} \le \epsilon
    \label{eq:max_fps} 
\end{equation}
where $fps(\sigma)$ and $a_1(\sigma)$ are the FPS and top-1 accuracy of design $\sigma$ respectively, $a_{1,\text{ref}}$ is the reference accuracy and $\epsilon$ is the user-specified maximum accuracy drop tolerance. An instance of such a use-case would be an AI Camera app that has to perform real-time scene detection over the high-resolution video captured by a smartphone, without catastrophically penalising accuracy.

\textbf{Use-case 2. TargetLatency:}
Optimise the application to achieve maximum accuracy while meeting a target latency constraint.
\vspace{-1mm}
\begin{equation}
    \small
    \max\limits_{\sigma} a_1(\sigma) \quad \text{s.t.} \quad T(\sigma) \le T_{\text{target}}
    \label{eq:target_latency} 
\end{equation}
where $a_1(\sigma)$ and $T(\sigma)$ are the top-1 accuracy and latency of design $\sigma$ and $T_{\text{target}}$ is the maximum tolerated latency defined by the user.
Such a scenario would be present when segmenting a speaker's video in order to apply AR features during video-conferencing. In this case, low response time, and thus latency, is critical in order to follow the movement of the user, while accuracy has to be maximised to provide meaningful outputs.

\textbf{Use-case 3. MaxAccMaxFPS:}
Optimise the application to achieve the maximum attainable accuracy and throughput.
\begin{equation}
    \small
    \label{eq:max_acc_max_fps}
    \max\limits_{\sigma} {\frac{a_1(\sigma)}{a_{1,\text{max}}} + w_{\text{fps}} \cdot \frac{fps(\sigma)}{fps_{\text{max}} }}
\end{equation}
where we set the accuracy weight to 1 and tune the FPS weight $w_{\text{fps}}$ to capture the relative importance between accuracy and throughput. As such, for equal importance, $w_{\text{fps}}$ can be set to 1.
Furthermore, the accuracy and FPS of each design $\sigma$ are divided by the maximum observed accuracy and FPS in the model space, to yield a non-dimensional objective function. 

For the use-cases in Eq.~(\ref{eq:max_fps}) and (\ref{eq:target_latency}), we reduce the MOO problems to a single objective by means of an $\epsilon$-constraint formulation~\cite{marler2004survey}. 
In the use-case of Eq.~(\ref{eq:max_acc_max_fps}), we adopt the weighted sum method~\cite{marler2004survey} and allow the user to specify an importance weight ($w_{\text{fps}}$) between the two metrics to be maximised. Overall, the flexibility of \tool's application model enables the expression of a wide spectrum of use-cases that capture the exact needs of a given DL application.

\textbf{Offline Optimisation.}
With accuracy being only a function of the reference model $m_{\text{ref}}$ and the applied transformation $t$, $a(\sigma)$ can be calculated offline on a user-supplied validation dataset for all values of {\small $t\in \mathcal{T}$}. On the other hand, with latency, throughput and memory depending on both the target device and the values of \tool's system parameters ($hw$), evaluating $T(\cdot)$, $fps(\cdot)$ and $mem(\cdot)$ is device-dependent and thus requires running each possible system configuration $\left<ce, N_{\text{threads}}, g, r \right>$ on the target device. The required values are obtained through the on-device runs performed by the \textit{Device Measurements} module (Fig.~\ref{fig:processing_flow}), which collects statistics, including min, max, average, median and $n$\textsuperscript{th} percentile of latency and throughput, together with peak memory usage.
Next, both the accuracy and device measurements are stored and organised in look-up tables.
As a final step, \tool considers the user-specified MOO problem and performs a complete enumerative search over the populated look-up tables, in order to yield the design $\sigma$ that optimises the given use-case.

\textbf{Run-time Adaptation.}
To sustain performance in spite of fluctuation in on-device resource availability, the \textit{Runtime Manager} applies an adaptation mechanism at run time. This involves re-tuning the configuration of the \textit{Application}, by selecting an alternative design that optimises the defined multi-objective function under the new conditions. To perform this step, the \textit{Runtime Manager} only stores the device-specific look-up tables.
At deployment time, the \textit{Application} regularly transmits system statistics, such as processor load, to the \textit{Runtime Manager}. In the event of a significant resource availability change (\textit{e.g.}~10\% difference in the GPU load), the \textit{Runtime Manager} is invoked as a separate thread, it searches the look-up tables for the new highest performing design and provides the \textit{Application} with the resulting configuration.

\begin{table}[t]
    \vspace{-0.7cm}
    \centering
    % \captionsetup{font=small,labelfont=bf}
    \caption{\small Target Platforms}
    \vspace{-0.2cm}
    \setlength{\tabcolsep}{2pt}
    \resizebox{\linewidth}{!}{
        % \scriptsize
        \begin{tabu}{l l l l}
            \toprule
            \textbf{Device} & 
            % \scriptsize \textbf{Model} &
            \textbf{Sony Xperia C5 Ultra} & 
            \textbf{Samsung A71} & 
            \textbf{Samsung S20 FE} \\
            \midrule
            Year & 2015, August & 2020, January & 2020, October \\
            
            Chipset
            & \begin{tabular}[l]{@{}l@{}} MediaTek MT6752 \end{tabular} & Snapdragon 730 & Exynos 990 \\
            
            \begin{tabular}[l]{@{}l@{}} CPU \\ \phantom{0} \\ \phantom{0} \\ \phantom{0} \\ \phantom{0} \\ \phantom{0} \end{tabular} & \begin{tabular}[l]{@{}l@{}} 8$\times$ 1.69 GHz \\ ARM Cortex-A53 \\ \phantom{0} \\ \phantom{0} \\ \phantom{0} \\ \phantom{0} \end{tabular} & \begin{tabular}[l]{@{}l@{}} 2$\times$ 2.2 GHz \\  Kryo 470 Gold \\ 6$\times$ 1.8 GHz \\  Kryo 470 Silver \\ \phantom{0} \\ \phantom{0} \end{tabular} & \begin{tabular}[l]{@{}l@{}} 2$\times$ 2.73 GHz \\ Exynos M5 \\ 2$\times$ 2.5 GHz \\ ARM Cortex-A76 \\ 4$\times$ 2.0 GHz \\ ARM Cortex-A55 \end{tabular} \\
            
            \begin{tabular}[l]{@{}l@{}} GPU \end{tabular} & \begin{tabular}[l]{@{}l@{}} Mali-T760 MP2 \end{tabular} & \begin{tabular}[l]{@{}l@{}} Adreno 618 \end{tabular} & \begin{tabular}[l]{@{}l@{}} Mali-G77 MP11 \end{tabular} \\
            
            \begin{tabular}[l]{@{}l@{}} NPU \end{tabular} & \xmark & \cmark & \cmark \\
            
            % L2 Cache & 512 KB & 128 KB & \stelios{Let's omit this row. It doesn't add much since we don't optimise for it.} \\
            
            RAM & 2 GB @ 800 MHz & 6 GB @ 1866 MHz & 6 GB @ 2750 MHz \\
            
            Android & 6.0 (API level 23) & 10 (API level 29) & 11 (API level 30) \\
            
            \begin{tabular}[l]{@{}l@{}} Camera API \end{tabular} & \begin{tabular}[l]{@{}l@{}} \texttt{LEGACY} \end{tabular} & \begin{tabular}[l]{@{}l@{}} \texttt{LEVEL\_3} \end{tabular} & \begin{tabular}[l]{@{}l@{}} \texttt{FULL} \end{tabular} \\
            Battery & 2930 mAh & 4500 mAh & 4500 mAh \\
            \bottomrule
        \end{tabu}
    }
    \vspace{-0.6cm}
    \label{tab:devices}
\end{table}

\subsection{Implementation}
\label{sec:implementation}
We implement \tool's offline component using \mbox{TensorFlow}~(v2.4.0). TensorFlow is used for building the \textit{Transformations} module (Fig.~\ref{fig:processing_flow}) and for evaluating the accuracy of the generated model variants. 
Specifically, we utilise TFLite's post-training quantisation mode. For the FP16 models, we use \texttt{float16} quantisation, while for the INT8 models we use TFLite's dynamic range quantisation mode.

\tool's mobile \textit{Application} and \textit{Runtime Manager} were developed in Java (v8). For optimised on-device inference, we utilise TFLite's interpreter, supporting CPU, GPU and NNAPI backends. 
To support a wide range of computer vision applications, we integrate into \tool's SIL layer Android's \texttt{Camera2} API, which gives full control over the device's camera sensors, and the \texttt{Room} library, which provides a local Gallery database for storing each user's labelled photos.

\vspace{-0.4em}
\section{Evaluation}
\label{sec:eval}
% \vspace{-0.1cm}

\vspace{-0.3em}
\subsection{Experimental Setup}
\label{sec:exp_setup}
\vspace{-0.2em}
We target three smartphones with different resource \mbox{characteristics} (Table~\ref{tab:devices}): the low-end Sony Xperia C5 Ultra, the mid-tier Samsung A71 and the high-end Samsung S20 FE. The evaluated devices also differ in terms of camera capabilities and Android API levels.
Each experiment is run 200 times, with 15 warm-up runs, to obtain the average latency.

\begin{figure}[t!]
    \centering
    \vspace{-0.5cm}
    \includegraphics[width=1.0\columnwidth,trim={0.25cm 0cm 0.25cm 0cm},clip]{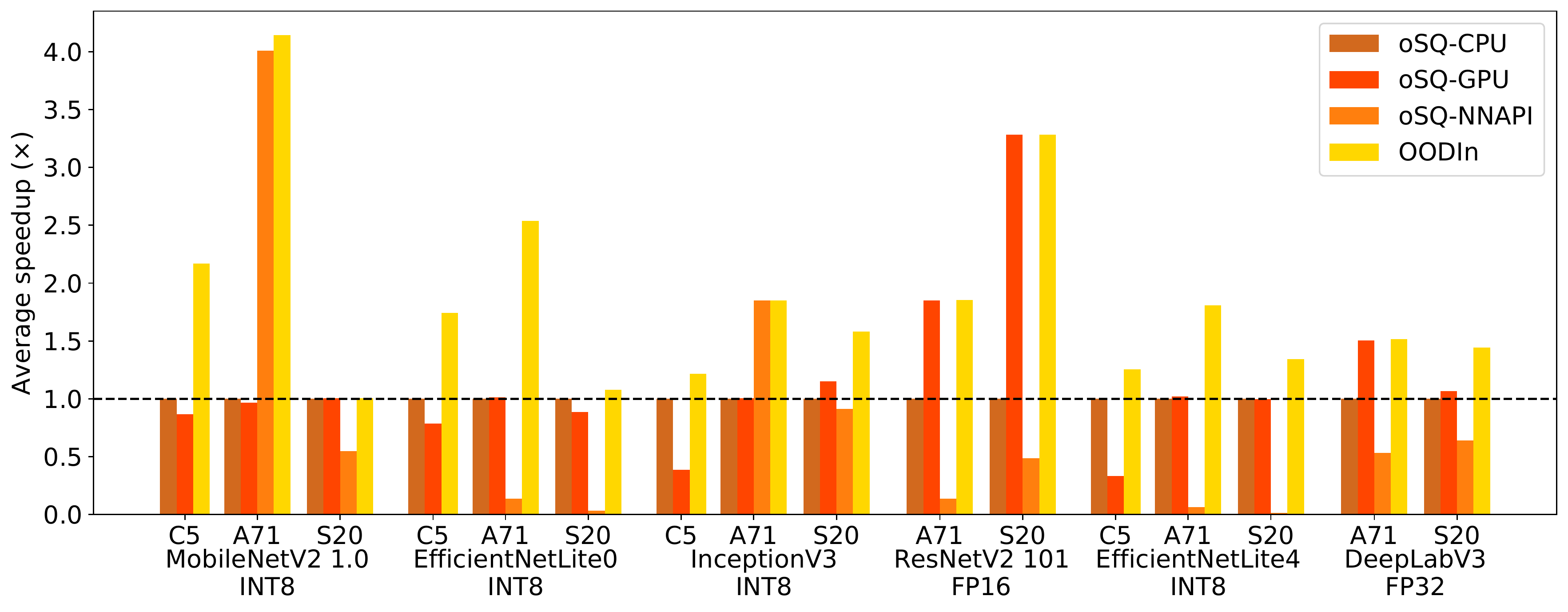}
    \vspace{-0.7cm}
    \caption{Comparison with optimised status-quo (oSQ-CPU, -GPU, -NNAPI) designs across devices and models. oSQ-CPU is used as the speedup baseline.}
    \vspace{-0.4cm}
    \label{fig:osqd}
\end{figure}

\begin{table}[t]
    % \vspace{-0.7cm}
    \centering
    % \captionsetup{font=small,labelfont=bf}
    \caption{\small Evaluated Deep Neural Networks}
    \vspace{-0.2cm}
    \setlength{\tabcolsep}{2pt}
    % \renewcommand{\arraystretch}{0.5}
    % \resizebox{\columnwidth}{!}{
    % \begin{threeparttable}
        \scriptsize
        \begin{tabu}{l c c c r c r}
            \toprule
            \textbf{DNN} & 
            \textbf{Precision*} & 
            \textbf{Resolution} &
            \textbf{Top-1/mIoU} &
            \textbf{Params} &
            \textbf{Size} &
            \textbf{FLOPs} \\
            \midrule
            MobileNetV2 1.0 & INT8 & 224$\times$224 & 70.8\% & 3.47 M & 3.41 MB & 0.6 G \\
            MobileNetV2 1.0 & FP32 & 224$\times$224 & 71.8\% & 3.47 M & 13.3 MB & 0.6 G \\
            EfficientNetLite0 & INT8 & 224$\times$224 & 74.4\% & 4.7 M & 5.17 MB & 0.8 G \\
            MobileNetV2 1.4 & FP32 & 224$\times$224 & 75.0\% & 6.06 M & 23.2 MB & 1.1 G \\
            EfficientNetLite0 & FP32 & 224$\times$224 & 75.1\% & 4.7 M & 17.7 MB & 0.8 G \\
            ResNetV2 101 & FP32 & 299$\times$299 & 76.8\% & 44.5 M & 170 MB & 15.6 G \\
            InceptionV3 & INT8 & 299$\times$299 & 77.5\% & 23.9 M & 22.8 MB & 11.4 G \\
            InceptionV3 & FP32 & 299$\times$299 & 77.9\% & 23.9 M & 90.9 MB & 11.4 G \\
            EfficientNetLite4 & INT8 & 300$\times$300 & 80.2\% & 13.0 M & 14.3 MB & 5.2 G \\
            EfficientNetLite4 & FP32 & 300$\times$300 & 81.5\% & 13.0 M & 49.4 MB & 5.2 G \\
            DeepLabV3 & FP32 & 513$\times$513 & 71.8\%%71.83\%(test)
            & 5.75 M & 2.65 MB & 5.7 G \\
            \bottomrule
            \end{tabu}
            \begin{tablenotes}
    				\scriptsize
    				\item *We omit FP16 precision from the table as it yielded accuracy within 1\% of FP32's.
    		\end{tablenotes}
        % \end{threeparttable}
    % }
    \vspace{-0.6cm}
    \label{tab:dnns}
\end{table}

\textbf{DNN Models.} To show the generalisability of \tool across models, we select five representative DNNs (Table~\ref{tab:dnns}) of varying depth, architecture and computational footprint: 1)~MobileNetV2, 2)~ResNetV2 101, 3)~InceptionV3, 4)~EfficientNet and 5)~DeepLabV3. MobileNetV2~\cite{sandler2018mobilenetv2} is a hand-crafted model for resource-constrained devices. We target two variants of increasing size (1.0 and 1.4). ResNetV2 101~\cite{resnet2016cvpr} and InceptionV3~\cite{inception2017aaai} are mainstream models that achieve high accuracy at the expense of significant resource demands. EfficientNet~\cite{efficientnet2019icml} is a state-of-the-art automatically generated DNN for mobile devices. We target two variants of increasing size (Lite0 and Lite4). Finally, DeepLabV3~\cite{deeplabv32017arxiv} is a state-of-the-art semantic segmentation model, used for image understanding applications. We use the mobile-friendly DeepLabV3 variant with a MobileNetV2 backbone pretrained on ImageNet and a depth multiplier of 0.5.  

% \noindent
\textbf{Tasks and Datasets.}
To demonstrate the applicability of \tool across tasks, we target two tasks: 1000-class image classification and 21-class semantic segmentation.
The image classification models were trained on ImageNet ILSVRC 2012 and we report accuracy on the 50k-images validation set. The DeepLabV3 segmentation model was first pretrained on the MS COCO dataset and then trained on PASCAL VOC 2012. We report the VOC test accuracy in terms of pixel intersectio-over-union (IoU) averaged across the 21 classes~\cite{deeplabv32017arxiv}.

\textbf{Baselines.} To assess \tool's performance against the state-of-the-art (SOTA), we compare with the following baselines: i)~optimised status-quo (oSQ-D), ii)~platform-aware (PAW-D) and iii)~model-aware (MAW-D) designs.

oSQ-D consists of conventional but optimised execution on a single compute engine. As such, we report results using CPU-, GPU- or NNAPI-only. For each compute engine, we tune the associated parameters to obtain an optimised implementation. In particular, for oSQ-CPU, we enable the use of the \texttt{XNNPACK} library and tune the number of threads between 1 and the device's $N_{\text{cores}}$. oSQ-CPU is equivalent to the SOTA CPU-based design in~\cite{integer_dnn2018cvpr}. 
For oSQ-GPU, we use the fastest between FP16 and INT8. 
oSQ-GPU is equivalent to a set of SOTA GPU-based designs~\cite{cnndroid2016mm0,deepmon2017mobisys,heimdall2020mobicom}. For oSQ-NNAPI, we use the default accelerator specified by the device vendor.

\begin{figure}[t!]
    \centering
    \vspace{-0.5cm}
    \includegraphics[width=1.0\columnwidth,trim={0.25cm 0cm 0.25cm 0cm},clip]{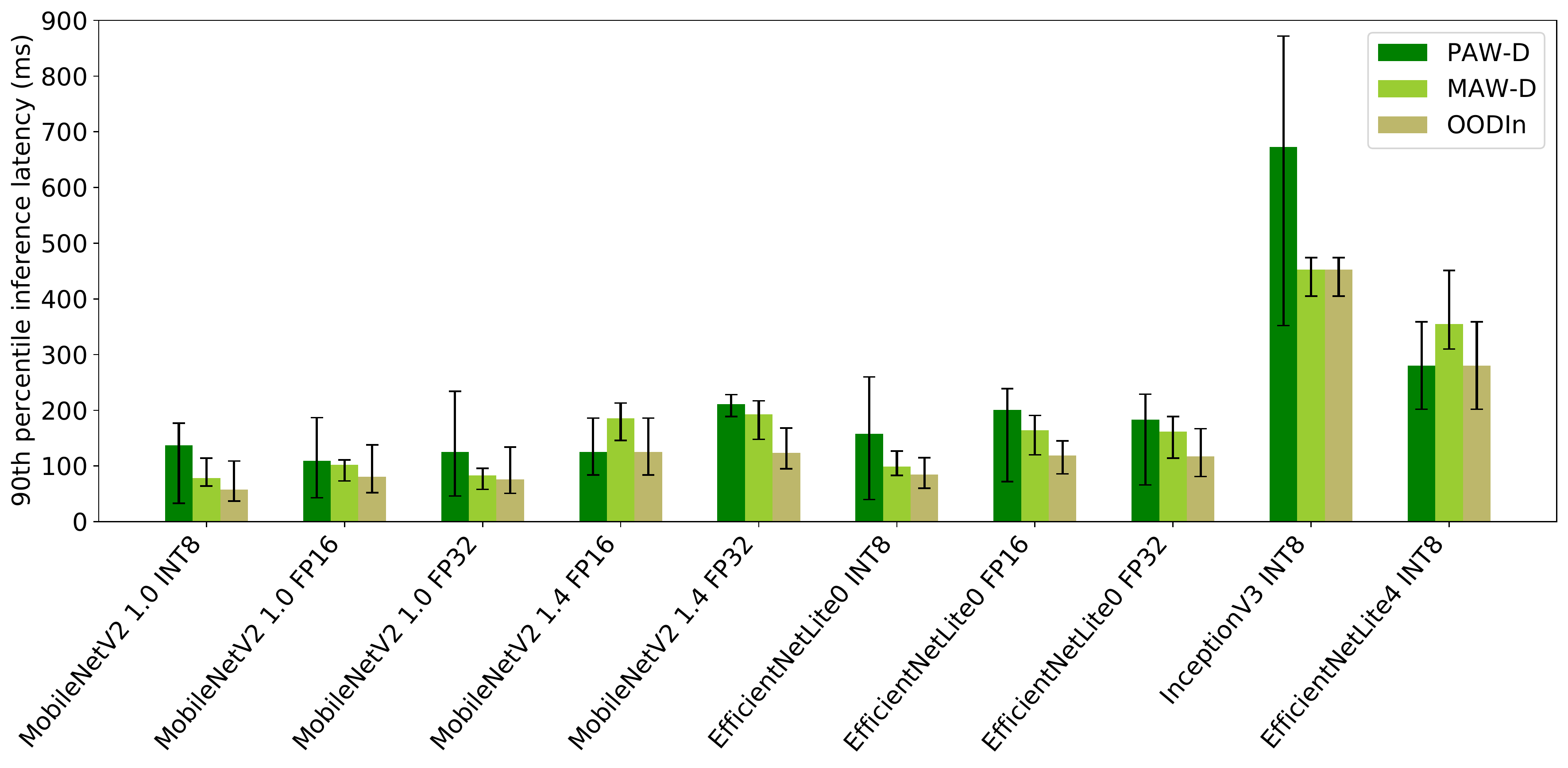}
    \vspace{-0.8cm}
    \caption{Comparison with PAW-D and MAW-D designs on low-end Sony Xperia C5 Ultra across various models. A subset of models are not depicted. These are DNNs that caused either thermal issues due to rapid overheating, or significant lag ($\ge 5$ s) of the AI Camera app, which would be catastrophic for the user experience and hence are not deployable on the particular device.}
    \vspace{-0.4cm}
    \label{fig:pawd_mawd_c5}
\end{figure}

\begin{figure}[t!]
    \centering
    \includegraphics[width=1.0\columnwidth,trim={0.25cm 0cm 0.25cm 0cm},clip]{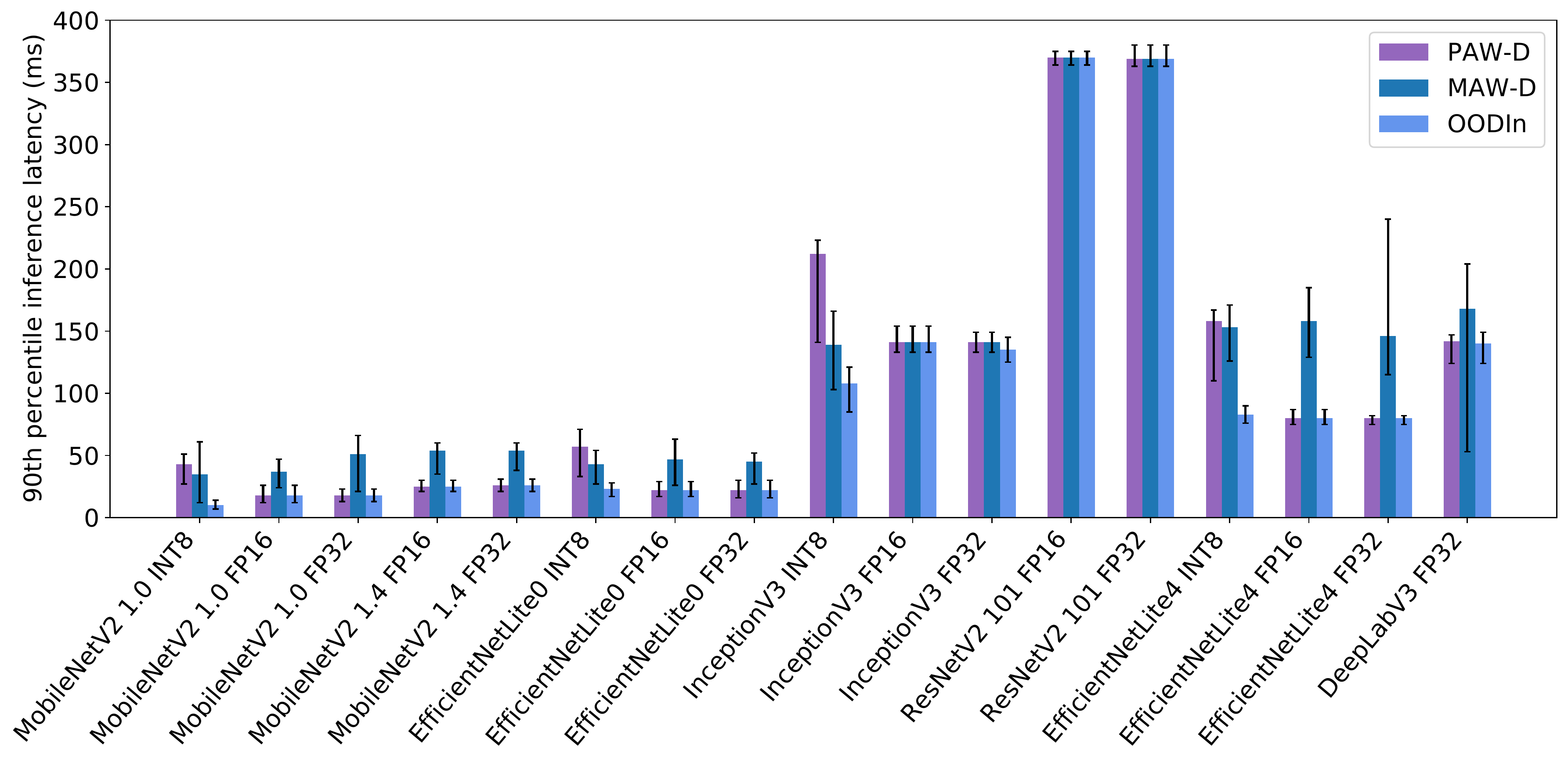}
    \vspace{-0.8cm}
    \caption{Comparison with PAW-D and MAW-D designs on mid-tier Samsung A71 across various models.}
    \vspace{-0.4cm}
    \label{fig:pawd_mawd_a71}
\end{figure}

\begin{figure}[t!]
    \centering
    \includegraphics[width=1.0\columnwidth,trim={0.25cm 0cm 0.25cm 0cm},clip]{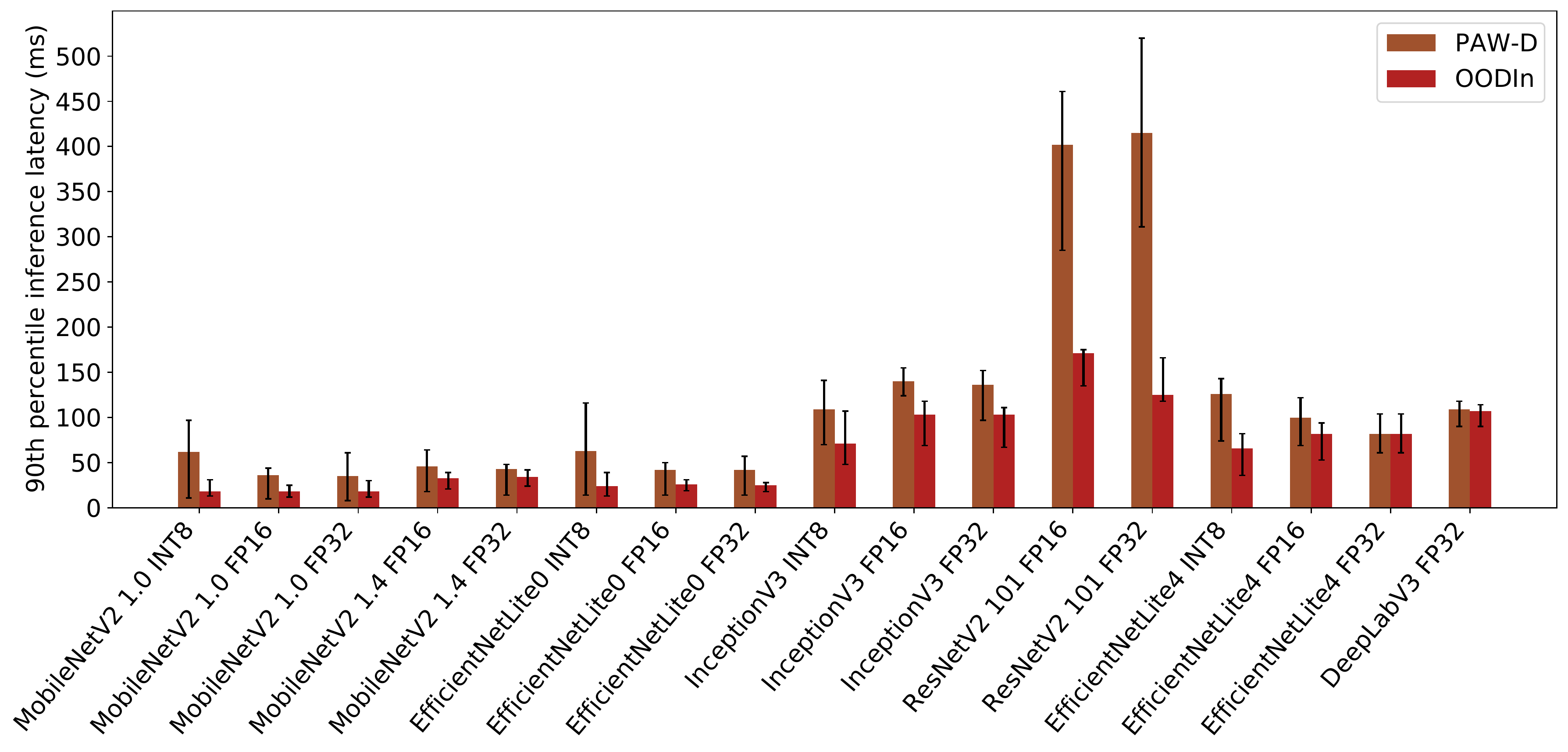}
    \vspace{-0.8cm}
    \caption{Comparison with PAW-D designs on high-end Samsung S20 FE across various models. We omit MAW-D designs as these have been optimised for S20 and hence coincide with \tool's designs.}
    \vspace{-0.6cm}
    \label{fig:pawd_s20}
\end{figure}

PAW-D comprises a \textit{model-unaware} configuration that has been optimised for the target device. To this end, we use EfficientNetLite4\footnote{We select EfficientNetLite4 as it lies in the middle in terms of computational and memory demands among the evaluated DNNs (Table~\ref{tab:dnns}).} for configuration optimisation on each device and use the \textit{same} configuration \textit{across models} for execution on the same device. \stelios{TODO - add correspondence to existing papers.}
MAW-D is a \textit{model-aware} configuration, but \textit{platform-agnostic}. As such, we use the target model for each experiment and optimise its configuration using a single device. We use S20 for this step, emulating the industry common practice of optimising for flagship devices. The resulting configuration is used \textit{across devices}. \stelios{TODO - add correspondence to existing papers.}

% \subsection{Evaluation of \tool's Design Space}
% \label{sec:eval_design_space}

% \begin{figure*}[t!]
%     \centering
%     \includegraphics[width=0.45\textwidth]{figures/latency_accuracy.pdf}
%     \caption{}
%     \label{fig:latency_accuracy}
% \end{figure*}

\vspace{-0.3em}
\subsection{Evaluation of \tool's Performance}
% \subsection{Performance Comparison}
\label{sec:eval_perf}
\vspace{-0.1em}

% This section assesses the actual performance gains of \tool over our optimised baselines.

\textbf{Comparison with oSQ-D.}
This section presents a comparison of our framework with optimised status-quo baselines. This is investigated by setting the objective of \tool to minimising the average latency with no accuracy drop allowed, and comparing the performance of the resulting configuration across devices and models.
Fig.~\ref{fig:osqd} shows the performance benefits achieved by \tool. Our framework delivers a speedup of up to 4.14$\times$ (1.73$\times$ geometric mean across the models), 4.29$\times$ (1.74$\times$ geo. mean) and 93.46$\times$ (5.9$\times$ geo. mean) over oSQ-CPU, oSQ-GPU and oSQ-NNAPI, respectively. Morevover, \tool provides 3.3$\times$ higher speedup over oSQ-NNAPI, compared to oSQ-CPU and -GPU. This indicates that, despite its potential, NNAPI remains in its infancy and naively resorting to NNAPI designs often leads to suboptimal execution.

We further observe that the highest performing engine changes as a function of both model and device. This phenomenon is rooted in the wide disparity between DNN workloads, but also highlights the vast system heterogeneity across devices that directly affects performance. 
Hence, the different oSQ-D designs become suboptimal as we move to a different DNN-device pair. 
On the other hand, \tool counteracts the effect of both DNN diversity and system heterogeneity and sustains high performance, by using its model- and system-level parameters to tailor the execution to the traits of both model and device.
Overall, the results demonstrate the performance benefits provided by \tool over CPU, GPU and NNAPI, as well as its scalability across diverse mobile devices.

% oSQ-CPU: range 1.01$\times$-4.14$\times$ (1.73$\times$ geometric mean across the models).

% oSQ-GPU: range 1$\times$-4.29$\times$ (1.74$\times$ geometric mean across the models).

% oSQ-NNAPI: range 1$\times$-93.46$\times$ (5.9$\times$ geometric mean across the models).

\textbf{Comparison with PAW-D and MAW-D.}
Here, we evaluate \tool with highly optimised platform- and model-aware designs. For these experiments, \tool's offline objectives were set to minimise the 90\textsuperscript{th} percentile inference latency subject to no accuracy drop with respect to the given DNN.

Fig.~\ref{fig:pawd_mawd_c5}-\ref{fig:pawd_s20} show the achieved performance across models for the low-, mid- and high-end devices, respectively.
% n Sony C5, a subset of models are not depicted. These are the DNNs that caused either thermal issues due to rapid overheating of the device, or significant lag ($\ge 5$ s) of the AI Camera application, which would be catastrophic for the user experience and hence are not deployable on the particular device.
Compared to PAW-D and MAW-D on Sony C5, we observe a speedup of up to 2.36$\times$ (1.49$\times$ geometric mean across the models) and 1.56$\times$ (1.30$\times$ geo. mean), respectively.
On Samsung A71, \tool yields performance improvements of up to 4.3$\times$ (1.25$\times$ geo. mean) over PAW-D and 3.5$\times$ (1.67$\times$ geo. mean) over MAW-D. 
Finally, on Samsung S20, our framework outperforms PAW-D by up to 3.44$\times$ (1.7$\times$ geo. mean).

% \stelios{TODO - Comment on the similar performance for DeepLabV3-A71-PAW-D and -S20, ResNet101-A71 and InceptionV3-C5.}

The performance gains come mainly from the fact that the proposed framework tailors the execution to both the model and platform characteristics. PAW-D has to rely on the traits of a proxy DNN to determine the execution configuration. 
% \stelios{Example - InceptionV3 on C5 is quite suboptimal. Most probably, PAW-D selected CPU (judging from EfficientNetLite4 Fig. 3 where CPU is the best), while GPU is the best for InceptionV3 on C5.} 
The suboptimality of this approach can be observed by examining the mapping of InceptionV3 on A71. In this case, PAW-D selects a GPU design, as this engine yielded the highest performance for the proxy DNN (EfficientNetLite4). On the other hand, \tool deploys the model on A71's NNAPI, which is the actual best configuration (Fig.~\ref{fig:osqd}), delivering 1.87$\times$ lower latency.
Similarly, although MAW-D considers the actual DNN, it optimises based on the behaviour of S20 and hence does not capture the hardware peculiarities of the actual target device.
The drawback of this is manifested for MobileNetV2 1.0 INT8. On S20, the CPU is often the highest performing engine (Fig.~\ref{fig:osqd}). As such, MAW-D opts in to use a CPU-based design for MobileNetV2 1.0 INT8 on A71. \tool, instead, selects an NNAPI-based design, achieving 3.5$\times$ speedup over MAW-D.
% \stelios{Example - CPU is the often the best for S20. See EfficientLite0-INT8. MAW-D selects the CPU based on S20's behaviour. However, the GPU would yield the best results on C5.} 
% As such, both lead to either suboptimal execution schemes or leave the device resources underutilised. \tool overcomes both limitations ... \stelios{Give more insight here on the "why".}
As such, our framework overcomes the limitations of both baselines by tailoring execution to the target DNN-device pair and thus better utilising the available device resources.

\begin{figure}[t!]
    \vspace{-0.3cm}
    \centering
    \includegraphics[width=0.48\textwidth]{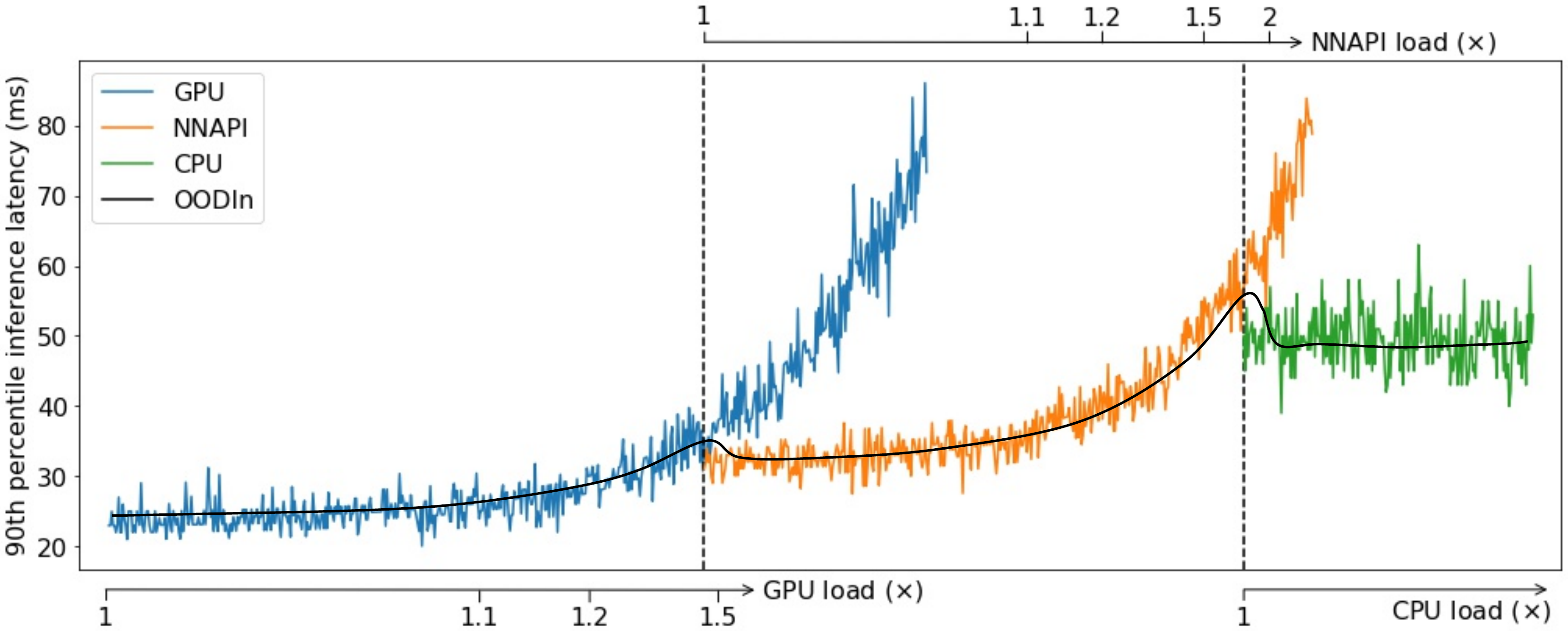}
    \vspace{-0.2cm}
    \caption{\textit{Runtime Manager}'s behaviour under device load, when targeting MobileNetV2 1.4 on A71.}
    \vspace{-0.2cm}
    \label{fig:rm_load}
\end{figure}

\begin{figure}[t!]
    \centering
    \includegraphics[width=0.48\textwidth]{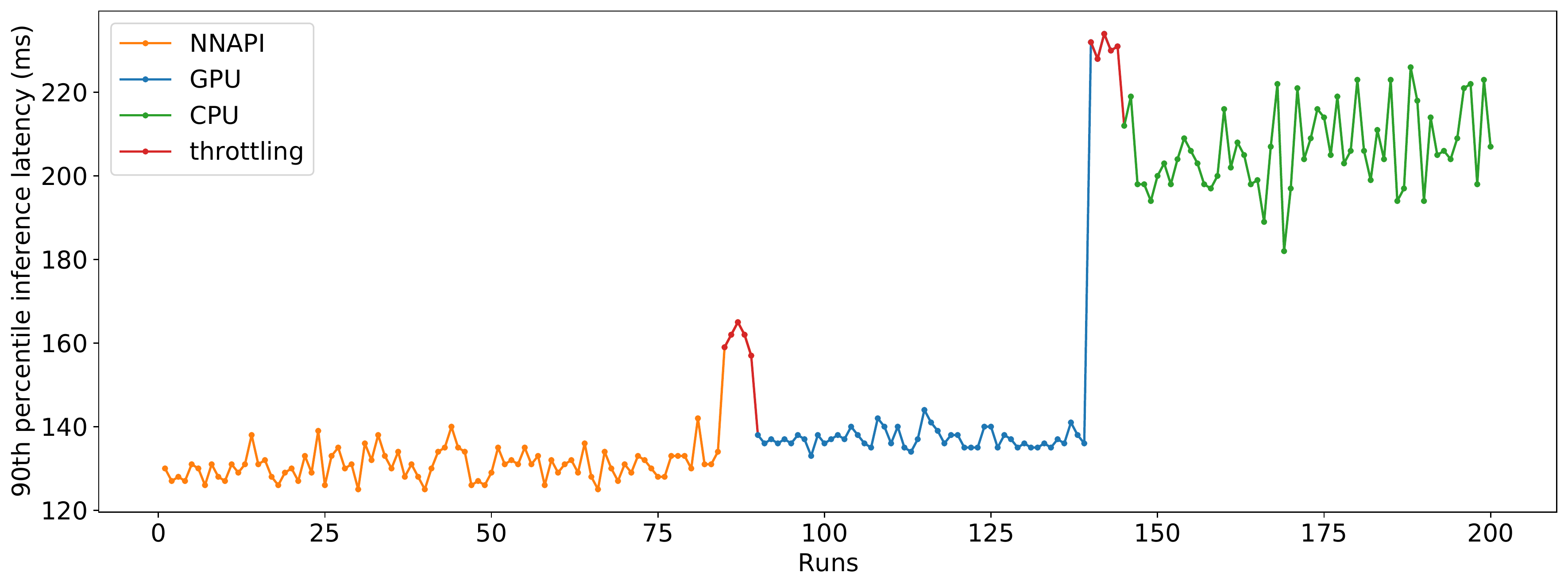}
    \vspace{-0.3cm}
    \caption{\textit{Runtime Manager}'s behaviour under thermal throttling, targeting InceptionV3 on A71.}
    \vspace{-0.6cm}
    \label{fig:rm_throttling}
\end{figure}

\vspace{-0.4em}
\subsection{Run-time Adaptation}
\label{sec:eval_runtime}
\vspace{-0.2em}

To assess the responsiveness of \tool's \textit{Runtime Manager} in adapting to dynamic changes, we targeted A71 and investigated \tool's performance under two characteristic scenarios: \textit{i)}~device load and \textit{ii)}~processor thermal throttling.

\textbf{Device Load.}
To evaluate \tool's performance under various device loads, we measured the inference latency of our framework when varying the load of the target device using MobileNetV2 1.4. This is accomplished by exponentially scaling the inference latency by a load factor (\textit{i.e.}~a factor of 2 corresponds to 2$\times$ slower execution). Fig.~\ref{fig:rm_load} depicts the achieved 90\textsuperscript{th} percentile latency under various device loads.

With all engines idle (start of x-axis), \tool selects the highest performing design for the particular device-DNN pair, which runs on the GPU. As the GPU load increases (towards the right-hand side) due to sharing between multiple tasks, performance degrades, resulting in excessive latency at the first vertical line. At that point, \tool's \textit{Runtime Manager} performs a compute engine change and switches to an NNAPI-based configuration. In this manner, it is able to sustain higher performance despite the GPU overload. Further towards the right, as the NNAPI-targeted engine also becomes loaded, the \textit{Runtime Manager} switches to a CPU-based design, alleviating the severe impact of processor contention on latency.
As a result, by periodically monitoring resource usage and re-evaluating the current configuration, \tool's \textit{Runtime Manager} is able to adapt the execution based on the device load, leading to latency reductions of up to 2.7$\times$ (1.55$\times$ geometric mean across future runs) over the statically selected design.

\textbf{Thermal Throttling.}
To assess the adaptability of \tool with respect to thermal throttling events, we conducted a throughput-driven experiment: here, \tool uses InceptionV3 to process a continuous stream of images from the camera, leading to processor overheating and subsequent reduction of its performance through DVFS. Fig.~\ref{fig:rm_throttling} shows the achieved inference latency as a function of inference runs.

Initially, \tool selects the highest performing configuration for the target device-DNN pair, which is NNAPI-based. 
After the 85\textsuperscript{th} processed image, the NNAPI's performance rapidly deteriorates due to frequency reduction as dictated by the DVFS governor. Our \textit{Runtime Manager} detects the throttling event within approximately 800 ms and switches to the next highest performing design that maps on a different engine, \textit{i.e.}~a GPU-based design for the particular device-DNN pair. After several inferences, GPU throttling is launched. \tool's \textit{Runtime Manager} detects it within approximately 1150 ms and switches to a CPU-based configuration. Overall, by identifying throttling behaviour in a timely manner, \tool's run-time adaptation mechanism effectively decides when to impose a new configuration and which compute engine to migrate to, providing the best performance possible at any given time instant independently of the system dynamicity.

\vspace{-0.2cm}
\section{Related work}
\label{sec:related_work}
\vspace{-0.1cm}

\textbf{Resource Characterisation for Mobile DL.}
\cite{rsc_charact2015iotapp} presented an initial study of the resource demands of mobile DL, investigating the energy consumption and processor and memory usage of DL applications, and how well these can be accommodated by mobile platforms.
Focusing on the wide diversity %in capabilities 
of existing devices, recent works have highlighted the impact of system heterogeneity when deploying DNNs~\cite{fb_edge2019hpca} and the significant performance variation between different platforms~\cite{embench2019emdl,ai_benchmark2019iccvw}. Moreover, \cite{dl_smartphones2019www} presents an empirical study on how real Android apps employ DL, showing that although 81\% of apps use DNNs, % for their core functionality, 
only 6\% of them apply optimisations.

\textbf{Mobile Backend Optimisations.}
With mobile chipsets increasingly hosting heterogeneous processors such as GPUs, DSPs and NPUs, prior work has investigated the optimised execution of DNNs in such settings. Focusing on single-processor execution, several frameworks~\cite{cnndroid2016mm0,deepmon2017mobisys,heimdall2020mobicom} have mapped the execution of DNNs on the GPU, while DeepEar~\cite{deepear2015ubicomp} also targets the DSP. 
Exploiting all available engines, another stream of work~\cite{ulayer2019eurosys,mobisr2019mobicom} has explored parallelising DNN inference across the heterogeneous engines.

\textbf{Adaptive Inference Systems.}
Aiming at adaptability to dynamic environments, \cite{neurosurgeon2017asplos} and \cite{dyno2021arxiv} introduced cloud-device collaboration by executing earlier layers locally and the rest on the cloud.
\cite{mcdnn2016mobisys} employs multiple models and dynamically decides which one and whether to execute it on device or cloud.
Recently, more specialised systems~\cite{hapi2020iccad,spinn2020mobicom,clio2020mobicom,mess2021arxiv} have been proposed, which exploit the different complexity of inputs to early-exit, resulting in faster processing and energy savings for simpler samples. Despite the high efficiency of these systems, their integration into mobile apps, as well as the generalisation of their use across heterogeneous devices, different DL models and applications remains an open problem.

\vspace{-0.2cm}
\section{Conclusion}
\label{sec:conclusion}

This paper presents \tool, a framework that mitigates the challenges of deploying high-performance DNNs across heterogeneous mobile devices.
By combining a novel mobile software architecture parametrised at both the model- and system-level, together with a multi-objective modelling framework for DL applications, \tool customises the execution to the target DNN, application needs and target device, outperforming both status-quo and platform- and model-aware designs across various smartphones. Moreover, through an efficient run-time adaptation mechanism, \tool dynamically reconfigures the execution to adapt to the inherent resource availability fluctuations of mobile devices, opening the way for robust and scalable on-device DL inference.

\bibliographystyle{./bibliography/IEEEtran}
\vspace{-0.2cm}
\bibliography{references.bib}

\end{document}